\documentclass[journal]{vgtc}                
\ifpdf
  \pdfoutput=1\relax                   
  
  \usepackage{graphicx}                
  \DeclareGraphicsExtensions{.pdf,.png,.jpg,.jpeg} 
\else
  \usepackage{graphicx}                
  \DeclareGraphicsExtensions{.eps}     
\fi%

\graphicspath{{figures/}{pictures/}{images/}{./}} 

\usepackage{microtype}                 
\PassOptionsToPackage{warn}{textcomp}  
\usepackage{textcomp}                  
\usepackage{mathptmx}                  
\usepackage{times}                     
\usepackage{cite}                      
\usepackage{tabu}                      
\usepackage{booktabs}                  
\usepackage{enumitem}
\usepackage{colortbl} 
\usepackage{verbatim}
\usepackage{amssymb}
\usepackage{bbm}
\usepackage{dsfont}
\usepackage{bbold}
\usepackage{mathbbol}
\usepackage{newtxmath}
\usepackage{threeparttable}     
\usepackage{multirow}
\usepackage{diagbox}
\usepackage{graphicx}
\usepackage{overpic}
\usepackage{gensymb}
\usepackage{url}
\usepackage{hyperref}
\usepackage[ruled,linesnumbered]{algorithm2e}

\usepackage{scrextend}
\deffootnote[0.5em]{0em}{1em}{\textsuperscript{\thefootnotemark}}

\usepackage{xurl}

\onlineid{1002}

\vgtccategory{Research}
\vgtcpapertype{Algorithm/Technique}

\title{Gaze-Vergence-Controlled See-Through Vision\\in Augmented Reality}

\author{Zhimin Wang*, Yuxin Zhao*, and Feng Lu$^{\dag}$, \textit{Senior Member, IEEE}}
\authorfooter{
\item *These two authors contributed equally to the paper.
\item $^{\dag}$Corresponding author: Feng Lu.
\item ZhiminWang and Yuxin Zhao are with the State Key Laboratory of Virtual Reality Technology and Systems, School of Computer Science and Engineering, Beihang University, Beijing 100191, China (e-mail: zm.wang@buaa.edu.cn; zyuxin@buaa.edu.cn).
\item Feng Lu is with the State Key Laboratory of Virtual Reality Technology and Systems, School of Computer Science and Engineering, Beihang University, Beijing 100191, China, also with the Peng Cheng Laboratory, Shenzhen 518055, China (e-mail: lufeng@buaa.edu.cn).
}

\shortauthortitle{Biv \MakeLowercase{\textit{et al.}}: Global Illumination for Fun and Profit}

\abstract{
	Augmented Reality (AR) see-through vision is an interesting research topic since it enables users to see through a wall and see the occluded objects. 
	Most existing research focuses on the visual effects of see-through vision, while the interaction method is less studied. 
	However, we argue that using common interaction modalities, \textit{e.g.}, midair click and speech, may not be the optimal way to control see-through vision.
	This is because when we want to see through something, it is physically related to our gaze depth/vergence and thus should be naturally controlled by the eyes.
	Following this idea, this paper proposes a novel gaze-vergence-controlled (GVC) see-through vision technique in AR.
	Since gaze depth is needed, we build a gaze tracking module with two infrared cameras and the corresponding algorithm and assemble it into the Microsoft HoloLens 2 to achieve gaze depth estimation.
	We then propose two different GVC modes for see-through vision to fit different scenarios.
	Extensive experimental results demonstrate that our gaze depth estimation is efficient and accurate.
	By comparing with conventional interaction modalities, our GVC techniques are also shown to be superior in terms of efficiency and more preferred by users.
	Finally, we present four example applications of gaze-vergence-controlled see-through vision.
	
} 


\CCScatlist{
	\CCScat{Augmented Reality}{See-through Vision}{Gaze Vergence Control}; Human Computer Interaction (HCI)
	
}

\teaser{
	\centering 
	\includegraphics[width=1\columnwidth]{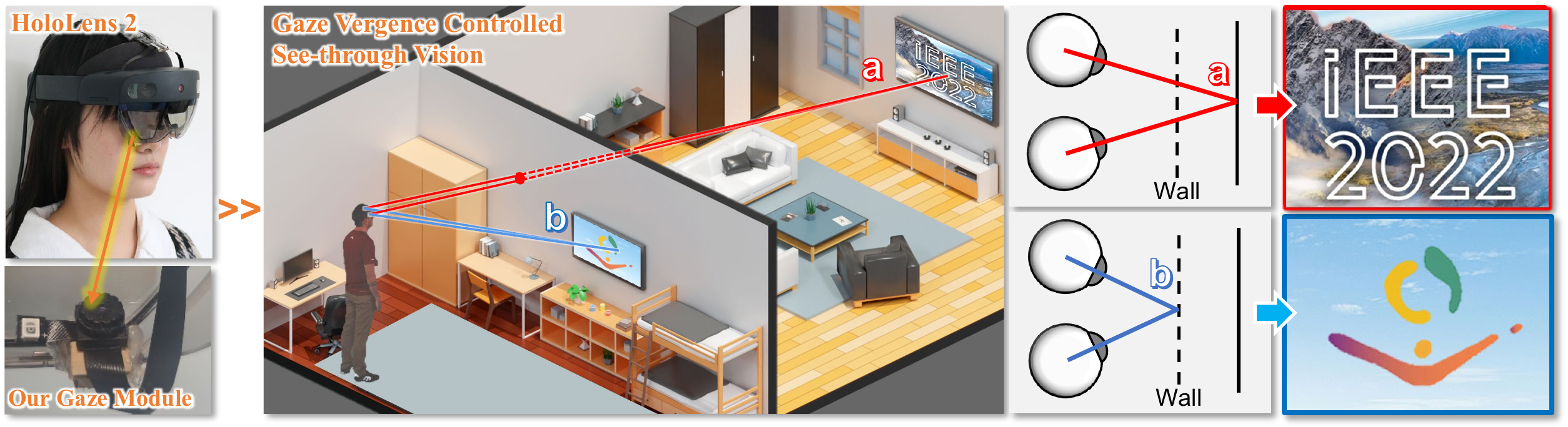}
	
	\caption{ We propose a gaze-vergence-controlled see-through vision in AR.
		We build a gaze tracking module with two infrared cameras and assemble it into the Microsoft HoloLens 2.
		With our gaze depth estimation algorithm, the user's gaze depth can be computed from gaze vergence and used to control see-through vision.
	}
	\label{figure:First-Image}
}

\vgtcinsertpkg

\begin{document}


\firstsection{Introduction}

\maketitle
Virtual Reality and Augmented Reality (VR \& AR) have attracted much attention from both academia and industry in the past five years.
In particular, with the rise of the meta-verse in recent years, AR and VR are widely considered the keys to the next generation of the internet \cite{metaverse}\cite{AWeighting}. 
The AR/VR industries continue to climb in market value \cite{AR_VR}.
These technologies are also utilized in a large number of applications from different fields, \textit{e.g.}, games, education and health care \cite{martinez2014drivers}.

While VR produces immersive virtual worlds generated by computer graphics, AR technology aims at enhancing the user experience by seamlessly integrating the virtual objects with the physical world \cite{KrevelenP10}.
The big tech giants, \textit{e.g.}, Microsoft and Apple, are also shifting their focus to AR and trying to apply AR technology in different areas such as intelligent manufacturing and online retail \cite{Apple}.

Since AR is able to link the real and virtual worlds,
one interesting application is to expand the user's vision, such as allowing the user to see the occluded objects behind a wall, namely see-through vision \cite{sandor2010augmented}.
The see-through vision has been explored in recent
years \cite{mori2017survey, swan2006perceptual, swan2007egocentric}.
Researchers have made efforts to improve the visual effect of see-through vision in AR \cite{DBLP:conf/ismar/AveryPT07, DBLP:conf/ismar/BarnumSDK09}. 
For instance, Avery \textit{et al.} designed the \textit{Edge Overlay} technique to provide depth cues for see-through vision \cite{DBLP:conf/vr/AveryST09}.
Erat \textit{et al.} presented the user's view with photorealistic rendering
from a three-dimensional reconstruction of hidden areas \cite{DBLP:journals/tvcg/EratIKS18}. 

The above works make see-through vision more natural and realistic.
However, the way to interact with see-through vision is less studied.
In fact, see-through vision can significantly benefit from interaction control, so as to enrich the user experience when using AR Head Mounted Display (HMD) devices \cite{billinghurst2009advanced, park2008wearable}.
By intention, the user can turn on/off see-through vision or show it at a different distance.
However, we argue that using the common interaction modalities, \textit{e.g.}, midair click and speech, may not be the optimal way to control see-through vision.
This is because when the user wants to see through a wall, he needs to think about the corresponding click gesture or speech command and then execute it.
It is not intuitive and requires extra effort to switch the thinking, which will distract the user's attention. 

Intuitively, the human eye gaze can be a more natural input to control see-through vision.
When we intend to see through something, we are actually fixating at a new distance, which is physically related to the gaze depth/vergence.
For instance, the gaze depth increases when we fixate on the occluded objects behind the wall, while it decreases when we look at the target at a nearer distance.


Inspired by this observation, a natural idea is to control see-through vision by gaze depth/vergence.
However, it is not easy to use gaze vergence to control see-through vision in AR HMDs.
The problems can be summarized in three aspects.
1) The mainstream AR devices do not support gaze depth estimation. 
For example, the Microsoft HoloLens 2 only offers single gaze ray but does not provide the gaze vergence or access to eye images \cite{Hololens_2}.
2) Recent studies using gaze vergence for interaction are mostly in desktop or VR scenarios \cite{DBLP:conf/chi/KudoOHSFK13, Verge}.
These methods rely on specific SDKs that cannot be easily adapted to the AR HMD.
3) There are few works in the literature that discuss how to flexibly control see-through vision by gaze depth.


%

To address these issues, our solution contains the following steps:
1) We build a gaze tracking module with two infrared cameras and assemble it into the Microsoft HoloLens 2, as shown in Fig. \ref{figure:First-Image}.
2) We design two gaze depth estimation methods, which can be easily adapted to different eye trackers. 
3) With our gaze depth estimation algorithm, we propose two control modes of gaze vergence and apply them to see-through vision.
We also investigate the efficiency of different modalities by quantitative performance measurements as well as subjective feedback.
Finally, we demonstrate the gaze-vergence-controlled techniques with four example applications{\footnote{Project page: \href{https://zhimin-wang.github.io/GVC_See_Through_Vision.html}{ https://zhimin-wang.github.io/GVC$\_$See$\_$Through$\_$Vision.html}}}.

Overall, our contributions are as follows:
\begin{enumerate}
	\item \textbf{Novelty}: We propose a Gaze-Vergence-Controlled (GVC) see-through vision technique in AR, offering new experiences.
	\item \textbf{System Implementation}: We customize two eye cameras and design gaze depth estimation methods for HoloLens 2.
	We also show that these methods are accurate and effective for see-through vision control.
	\item \textbf{Control Modes}: We propose two control modes of gaze vergence for see-through vision, which are called Stimulus-Guided (SG) see-through mode and Self-Control (SC) see-through mode.
	\item \textbf{Evaluation}: We demonstrate the efficiency and usability of our method through comparison and analysis.
	Four example applications of gaze vergence control are presented.
	
\end{enumerate}

\section{Related Work}

In this section, we review see-through vision and gaze interaction in AR and discuss the estimation methods of gaze depth. 

\subsection{See-Through Vision}

Occlusion visualization has been extensively explored in recent years. 
Elmqvist \textit{et al.} reviewed fifty techniques of occlusion management and classified them into five patterns \cite{elmqvist2008taxonomy}. 
We mainly concentrate on two patterns that are related to our work.

\textit{See-through vision.} 
The see-through vision can make the occluding surface partially transparent to turn objects visible \cite{mori2017survey, DBLP:journals/cga/JulierBBL02}.
Researchers have made efforts to improve the visual effect of see-through vision in AR \cite{DBLP:conf/ismar/AveryPT07, DBLP:conf/ismar/BarnumSDK09, DBLP:conf/ismar/KamedaTO04, DBLP:conf/mum/GruenefeldBB20}.
For instance, Avery \textit{et al.} provided see-through visualization with depth cues when users viewed hidden objects behind walls \cite{DBLP:conf/vr/AveryST09}. 
Erat \textit{et al.} synthesized three-dimensional models of occluded areas for presenting the photorealistic see-through vision. 
They also controlled a camera drone to explore the real scene via hand gestures and gaze direction \cite{DBLP:journals/tvcg/EratIKS18}.
Bane \textit{et al.} presented four interactive tools that allow users to explore see-through vision with different perspectives \cite{DBLP:conf/ismar/BaneH04}.

\textit{Multi-perspective visualization.} 
The multi-perspective vision is characterized by transforming an alternative view into the main window \cite{DBLP:journals/tvcg/YuZNDVG20, DBLP:journals/tvcg/WangWYP19}.
Prior studies captured occluded regions from  the secondary perspective and integrated them seamlessly into the user's view \cite{DBLP:journals/tvcg/WuP18, DBLP:journals/tvcg/VeasGKS12}. 
Lilija \textit{et al.} compared four different views for occluded object manipulation \cite{DBLP:conf/chi/LilijaPBH19}.
They found see-through vision had the best performance.

To summarize, previous literature mainly focused on the overlay effect of hidden areas and occluding layers.
However, the interaction method is less studied.
In fact, the see-through vision can significantly benefit from the interaction control.
According to the intention, the user can turn on/off see-through vision or show it at a different distance.
However, we argue that using the common interaction modalities, \textit{e.g.}, midair click and speech, may not be the optimal way to control the see-through vision.
This is because when we want to see through something, it is physically related
to our gaze depth/vergence and thus should be naturally controlled by the eyes.
Inspired by this fact, we propose a novel gaze-vergence-controlled see-through vision in AR.

\subsection{Gaze Interaction in AR}

Interaction techniques aim to improve the user experience, which is vital for AR HMD devices.
With the rise of gaze estimation accuracy \cite{9050633, DBLP:conf/ismar/WangZLL21, DBLP:conf/iccv/LiuLWL21}, different gaze-based techniques have been explored, such as gaze dwelling and vergence eye movement.

\textit{Gaze dwelling.} 
Most existing works exploit gaze dwelling  as the input technique \cite{wang2020comparing, DBLP:conf/chi/KytoEPLB18, DBLP:conf/huc/VidalBG13}. 
For instance, Wang \textit{et al.} used one second as the dwell time of selection for gaze-based interaction \cite{wang2021interaction}.
However, such gaze inputs often suffer from the Midas Touch problem \cite{DBLP:conf/ismar/MohanGFY18}, where users unintentionally trigger selections with natural eye movements.

\textit{Vergence eye movement.} 
Recent research tried to achieve Midas-touch-free interaction with vergence eye movement \cite{DBLP:conf/chi/KudoOHSFK13, DBLP:conf/chi/KirstB16, Verge, 10.1145/2634317.2634344, 10.1145/2678025.2701384, DBLP:conf/vr/WangZ022}.
For instance, Hirzle \textit{et al.} controlled the presentation of hidden virtual content triggered by gaze vergence \cite{DBLP:conf/chi/HirzleGGBR19}.
Compared with gaze dwelling, confirming selections via gaze vergence can be clearly distinguished from random visual skimming of the interface.
Therefore, vergence eye movement has the inherent advantage of addressing the Midas Touch problem.
However, there are few works in the literature that discuss how to flexibly control see-through vision by gaze depth.
To this end, we propose two control modes of gaze vergence
for see-through vision, which are called Stimulus-Guided (SG)
see-through mode and Self-Control (SC) see-through mode.


\subsection{Gaze Depth Estimation}

Many studies have investigated how to compute the gaze depth,
which can be broadly classified into two categories:
1) gaze ray-casting methods and 2) vergence-based methods.

\textit{Gaze ray-casting methods.} 
In these methods, the single gaze ray intersects the first object in the scene, and the intersection is taken as the 3D Point of Regard (PoR) \cite{DBLP:conf/sgda/MantiukBT11, DBLP:conf/etra/WeierRHS18}.
The distance between the PoR and the center of both eyes is defined as the gaze depth.
However, these methods do not deal with the occlusion ambiguity where multiple objects interact with the gaze ray, as they do not estimate the gaze depth directly.
Therefore, the gaze ray-casting methods are not suitable for the gaze-vergence-controlled technique.



\textit{Vergence-based methods.} The gaze vergence will change quickly when both eyes simultaneously move in opposite directions to fixate on objects at different depths.
The vergence-based methods generally include indirect and direct methods.
These indirect techniques first compute vergence-related features from near-eye images, \textit{e.g.}, Inter-pupillary Distance (IPD), and then use them to regress the gaze depth \cite{DBLP:conf/etra/MardanbegiCG19, pfeiffer2008evaluation, DBLP:journals/ki/OrloskyTSK16}. 
For instance, Alt \textit{et al.} detected the pupil diameter and IPD to estimate gaze depth and hence enabled gaze-based interaction with 3D virtual objects \cite{DBLP:conf/iui/AltSARB14}.
These direct methods obtain the gaze depth by computing the intersection of the gaze rays from both eyes \cite{DBLP:conf/chi/KirstB16, ETRA20}.
However, it is yet unclear as to which method could achieve better performance in AR HMD.
In this work, we implemented and compared two widely used methods in HoloLens 2, \textit{i.e.}, 3D line-of-sight intersection \cite{DBLP:conf/chi/KirstB16} and IPD-based regression \cite{kwon20063d, DBLP:conf/chi/KudoOHSFK13}.



\begin{figure*}[!t]
\centering 
\includegraphics[width=0.95\linewidth]{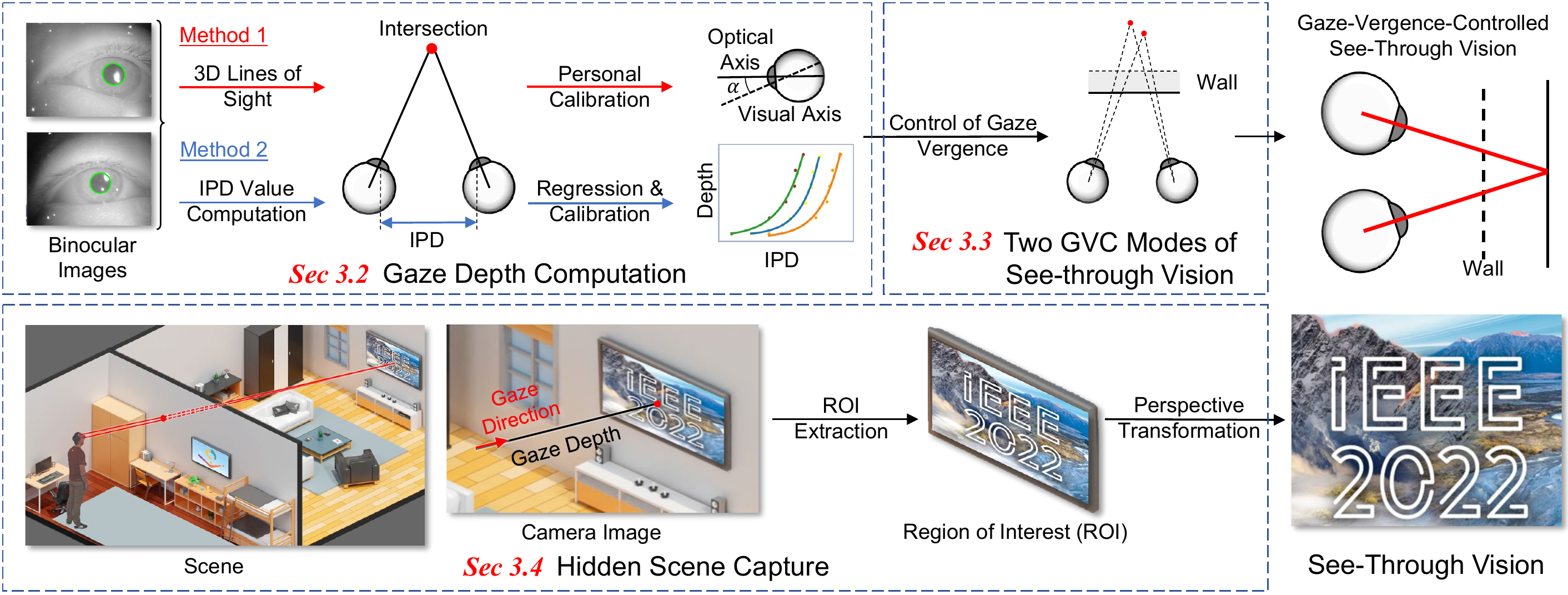}
\caption{Overview of our system. 
	We propose the gaze-vergence-controlled see-through vision technique in AR.
	We first get gaze depth from the proposed gaze depth estimation algorithm. The \textcolor[RGB]{255, 29, 29}{red} and \textcolor[RGB]{46, 117, 182}{blue} arrows indicate our 3D Line-of-sight Intersection and IPD-based Regression methods, respectively. With our algorithm, we then design two gaze-vergence-controlled modes of see-through vision. Finally, we capture the hidden scene using a camera behind the wall. The camera's view is seamlessly transformed into the user's view.
}
\label{figure:Overview}
\end{figure*}

\section{System Design}

\subsection{Overview}

We propose a novel gaze-vergence-controlled see-through vision in AR. 
An overview of our work is shown in Fig. \ref{figure:Overview}.
To compute the gaze depth/vergence, we first design the gaze depth computation module.
This module utilizes two methods to compute gaze depth, which are the 3D Line-of-sight Intersection (3D LosI) and the Inter-pupillary Distance (IPD) based regression. 
Based on the predicted depth, we further propose two gaze-vergence-controlled modes of see-through vision.
One is the Stimulus-Guided (SG) see-through mode and the other is the Self-Control (SC) see-through mode.
Besides the interaction techniques, we also introduce how to present a natural visual effect of see-through vision, which reveals the hidden scene.

The rest of this section is organized as follows.
1) We introduce the gaze depth computation in Section 3.2 including 3D line-of-sight intersection and IPD-based regression.
2) The two gaze-vergence-controlled modes of see-through vision are introduced in Section 3.3.
3) We describe the presentation of see-through vision from the hidden scene in Section 3.4.
4) We finally provide the implementation details of this system at the end.

\subsection{Gaze Depth Computation}

The gaze depth is defined as the distance between the user's PoR and the center of both eyes.
We can compute the PoR using the intersections of the lines of sight from the left and right eyes.
We build a gaze tracking module with two Near-Infrared (NIR) cameras and assemble it into the Microsoft HoloLens 2.
Here we utilize two gaze depth computation methods.
1) The first way is to directly compute the 3D intersections of the lines of sight, as described in Section 3.2.1.
2) The other way is an implicit model, which takes the IPD as input and regresses the gaze depth, as presented in Section 3.2.2.
The gaze vergence control combines the two methods, which will be described later in Section 4.

\subsubsection{Method 1: 3D Line-of-sight Intersection}

The 3D Line-of-sight Intersection (3D LosI) method uses the intersections of gaze rays from the left and right eyes.
Because HoloLens 2 only provides a single line of sight, we need to modify it to support binocular gaze estimation.
The mainstream strategy is to integrate HoloLens with the Pupil Labs' eye tracker and use its software \cite{COGAIN18, ETRA20}.
However, this method does not calibrate the combined hardware beforehand.
Instead, it merges the transformation between the scene camera and the eye camera with the kappa angle as a matrix to optimize.
The kappa angle is the angle offset between the optical and visual axes \cite{8451912}.
This way causes an increase in systematic error \cite{ETRA20}. 

\textbf{3D Lines of Sight}.
To improve computation accuracy, our method is modified from Pupil Labs' method in two ways:
1) employ the pupil detection method PuReST \cite{PuReST}, which has robust performance to reflections or partial occlusion;
2) calibrate the hardware in advance and model the kappa angle.
The goal of hardware calibration is to register the scene camera and eye cameras to a common coordinate system.
A more detailed description of our calibration procedure follows in Section 3.5.1.
The kappa angle is calculated by modeling the angle offset $\alpha$  between the visual and optical axes.
The explicit definition of the kappa angle helps to compensate for the estimation error.
We finally obtain two lines of sight from the left and right eyes.

\begin{figure}[t]
	\centering 
	\includegraphics[width=1\columnwidth]{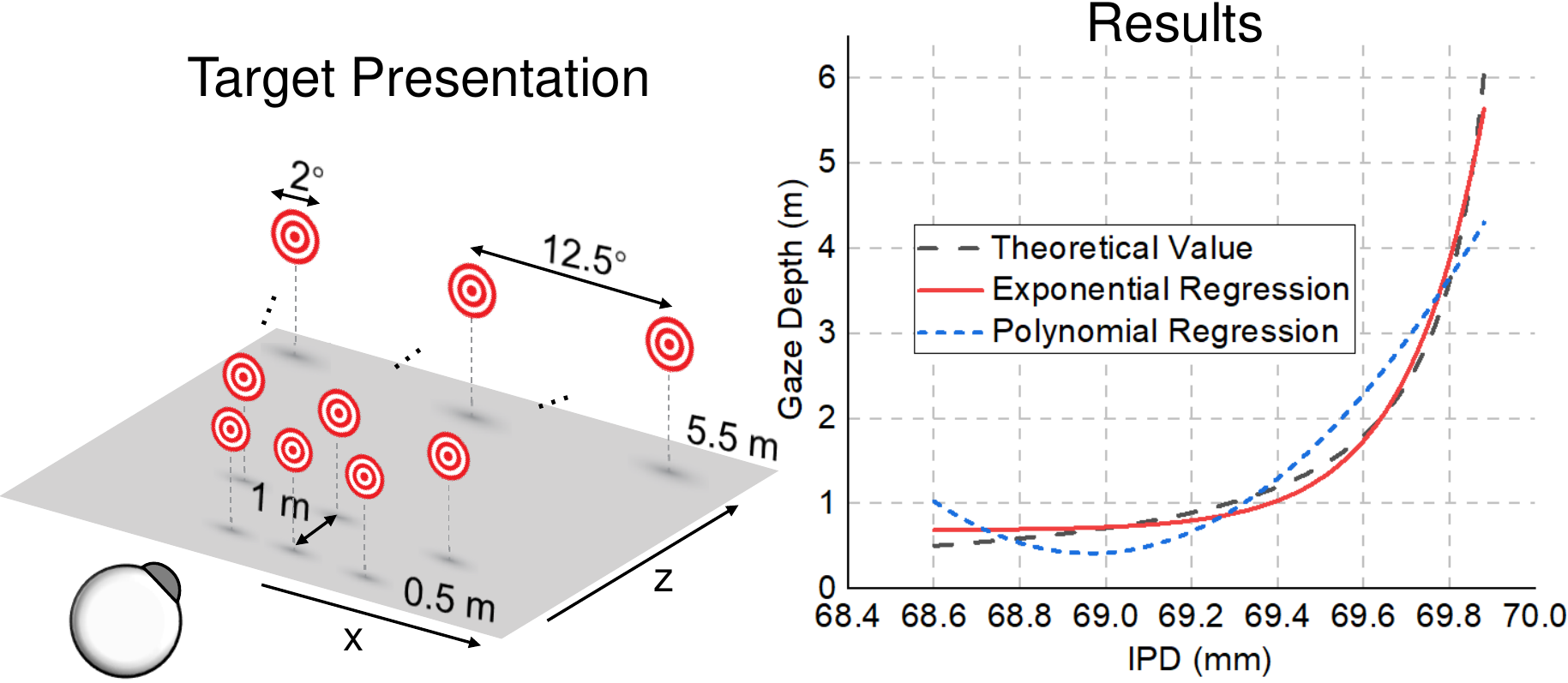}
	\caption{
		Depth calibration and fitting. Subjects view the calibration scene consisting of gaze targets that are distributed in depth (left). The comparisons among the simulation (dashed black line), the exponential regression (solid red line), and the polynomial regression (dashed blue line) of the theoretical relationship between gaze depth and IPD (right).
	}
	\label{figure:depth_fitting}
\end{figure}

\textbf{ Personal Calibration}.
We design a calibration scene to compute person-specific kappa angle  $\hat{\alpha}$, as shown in the left part of Fig. \ref{figure:depth_fitting}.
The gaze targets are displayed at depths between 0.5 m and 5.5 m and the distance interval in $z$ axis direction is 1 m. 
The duration of each point is 2 seconds and we only record data during the last second.
They are also scaled to subtend $2\degree$ of visual angle at all distances.
The movement directions of peripheral targets at $x$-$z$ plane keep 12.5° with the $z$ axis.
The y coordinates are set to the height of the user's head.
We collect some amounts of pupil data and gaze targets.
Finally, we apply a least squares algorithm to optimize the kappa angle as Chen \textit{et al.} did \cite{3D_Eye_Gaze}.

\textbf{3D Gaze Intersection}.
After the above two procedures, we obtain accurate binocular gaze rays. 
Then we can calculate the intersections of two gaze rays as the 3D gaze points, using the function denoted as equation (7.14) in \cite{swirski2015gaze}. 
The gaze depth is the distance from the center of both eyes to the 3D intersection point.

\subsubsection{Method 2: IPD-based Regression}

The 3D LosI highly relies on the accuracy of binocular 3D gaze estimation.
So it is also important to design a method that is insensitive to the line of sight. 
In this section, we introduce a technique that takes the IPD as input to regress gaze depth.
Specifically, we implement two IPD-based methods:
one utilizes the physical-based IPD in Millimeters (MIPD) to fit gaze depth, 
and the other uses the image-based IPD in Pixels (PIPD) to regress the depth.

Compared with previous IPD-based studies, our methods differ in some aspects.
First, prior research was mainly explored in the desktop environment or virtual reality settings \cite{ DBLP:conf/chi/KudoOHSFK13, DBLP:conf/iui/AltSARB14}, which cannot be easily adapted to the AR HMD.
Our module can be smoothly assembled with HoloLens 2.
Second, we employ the robust pupil detection and accurate eye model fitting method, which have been demonstrated with superiority to previous methods \cite{PuReST, dierkes2019fast}.
Another difference is the regression method.
Previous research used the support vector regression or the neural network to learn the mapping from IPD value to gaze depth \cite{DBLP:conf/egve/LeePELDB17, DBLP:conf/etra/WeierRHS18} while we theoretically verify that exponential regression is enough for the task.

\textbf{IPD Value Computation}. 
As indicated above, there are two ways described as follows.
We first perform the following procedure to obtain the MIPD: 
1) Building the physical models of both eyes.
We use the latest proposed 3D eye model fitting method \cite{dierkes2019fast}, which can mitigate the effects of corneal refraction and apply the two-sphere eye model.
2) Both eye models are registered to a common coordinate system according to the calibration parameters of hardware, as described later in Section 3.5.
We assume $p_{l}$ and $p_{r}$ are the 3D pupil centers of the left and right eyes.
The MIPD is estimated as 
$\theta_1=\left\|p_{l}-p_{r}\right\|_1$.  
We then compute the PIPD from each pair of eye images.
Let $x_{l}$ and $x_{r}$ be the horizontal coordinates of the left and right pupil centers in images. 
The resolution of each image is  $320 \times 240$ pixels.
Therefore, the PIPD can be delivered as $\theta_2=320-x_{l}+x_{r} $.

\textbf{Regression for Depth}. IPD-based regression needs to build a mapping from IPD value to gaze depth.
To find an optimal mapping function, we simulate the relationship between gaze depth and IPD value theoretically, as the dashed black line shows in the right part of Fig. \ref{figure:depth_fitting}.
We set the distance between both eyeball centers as 70 mm and the radius of the eyeball as 10.39 mm, which are provided by Pupil Labs \cite{Pupil}.
According to the upward trend, we try to fit the simulation using polynomial and exponential regression.
The results demonstrate that the exponential fitting approximates the theoretical value.
It is similar to the finding by Kwon \textit{et al.} \cite{kwon20063d}, which used a logarithmic function to do that.
Our regression function can be written as
\begin{equation}
	\label{estimation_1}
	\hat{d}=k_{1} \cdot \exp \left(k_{2} \cdot \left(\theta-\bar{\theta}\right)\right)+k_{3}, \quad \bar{\theta}=\frac{1}{n} \sum_{i=1}^{n} \theta_{i},
\end{equation}
\begin{equation}
	\label{estimation_2}
	\hat{K}=\arg \min _{K}\left\{\sum_{i=1}^{n}\left(d_{i}-\hat{d}_{i}\right)^{2}\right\}, \quad K=\left\{k_{1}, k_{2}, k_{3}\right\}, 
\end{equation}
where $\hat{d}$ is the estimated depth value while $d$ is the truth value. 
$\theta$ is the IPD value, and its units can be millimeters or pixels.
The $\theta$ subtracts the average value $\bar{\theta}$ for accelerating the parameter fitting.
$n$ is the number of gaze targets collected in the calibration procedure.
$\hat{K}$ is the optimal parameter set.
We also combine the regression with the Random Sample Consensus (RANSAC) \cite{fischler1981random} to discard outliers.

\textbf{Personal Calibration}. We utilize the same calibration scene as in Section 3.2.1 to compute the parameters $\hat{K}$.
We split gaze targets into three sets according to horizontal FOV distributions and fit three exponential functions, respectively.
An example of exponential fitting is shown in the top center part of Fig. \ref{figure:Overview}.
In the prediction period, we divide horizontal FOV into three sections, which are consistent with three functions.
The gaze depth of each test datum is computed by the exponential function from the corresponding section.

\begin{figure}[!t]
	\centering 
	\includegraphics[width=0.8\columnwidth]{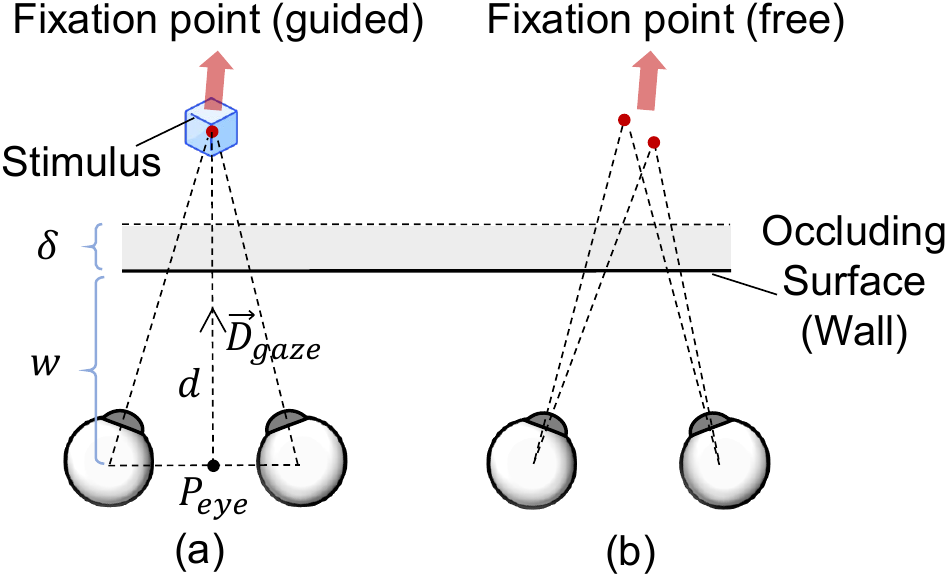}
	\caption{Two control modes of gaze vergence when a user looks towards the occluded surface. (a) Stimulus-guided See-through mode. (b) Self-control See-through mode.}
	\label{figure:control_mode}
\end{figure}

\subsection{Two Control Modes of See-through Vision}

Based on the two carefully designed gaze depth computation methods above,
we can successfully obtain the gaze depth.
But there is another equally important matter that is how to control see-through vision by gaze depth.
In this section, we design two different gaze-vergence-controlled modes. One is the Stimulus-Guided (SG) see-through mode, and the other is the Self-Control (SC) see-through mode.
The character of the first mode is simple and easy-to-use for users,
while the second mode is more novel and attractive.
We can thus choose a more suitable mode according to the specific application scenarios in AR.
To the best of our knowledge, there are few works in the literature that discuss how to flexibly control see-through vision by gaze depth. 

\textbf{Stimulus-Guided (SG) see-through mode}.
This mode allows users to trigger see-through vision by looking at a semi-transparent virtual stimulus behind the wall, as shown in Fig. \ref{figure:control_mode}a, which is similar to our viewing habit.
We attempt some variations of the representation with the purpose of the stimulus being as noninvasive as possible for the user, \textit{e.g.}, the size and transparency.
We finally choose a purple cube with a length of 10 cm and a transparency of 50$\%$.
This cube is attached to the user's gaze ray, which is located 6 meters away from the eyes.
The user first stands facing the wall.
Then the participant employs the stimulus as visual guidance, and thus the fixation depth increases for exceeding the threshold of activation.
Finally, the window of see-through vision is presented at a fixed distance, which helps to keep the PoRs fixated at a certain distance.
More formally, the window position $P_{\textit{\text{window}}}$ of see-through vision in $\mathbb{R}^{3}$ is calculated as
\begin{equation}
	\label{eqn_3}
	\gamma= \begin{cases}w+ j \cdot \delta, & \Phi \left(d\right)>w+\delta; \\ -\infty, & \text { otherwise, }\end{cases}
\end{equation}
\begin{equation}
	\label{eqn_4}
	P_{\textit{\text{window}}}=P_{\textit{\text{eye}}}+\gamma \cdot \vec{D}_{\textit{\text{gaze}}} ,
\end{equation}
where $\gamma$ is the window depth of see-through vision.
$w$ is the distance from the user to the wall, while $\delta$ is the distance threshold, as shown in Fig. \ref{figure:control_mode}a.
$j$ is a scale factor greater than 1.
$d$ is the estimated depth value, and
$\Phi\left(\cdot\right)$ is the filter function for data smoothing.
$P_{\textit{\text{eye}}}$ is the center of both eyes, and $\vec{D}_{\textit{\text{gaze}}}$ is the normal vector of the gaze ray.

In the above-proposed model, some parameters need to be determined.
A natural question arises: how to set reasonable parameters in practice.
We set the range of $w$ as (0.5, 3] m.
This is because this distance range is the most commonly used range for daily indoor interaction.
We call this distance range the middle distance in the context of our paper, referring to  Bardins \textit{et al.} \cite{10.1145/1461893.1461903}, 
To increase the robustness of control, the $\delta$ is determined by the mean and standard deviation of estimation error (see Section 4 for a more detailed description).
We use the mean filter as the $\Phi\left(\cdot\right)$ and the time window is empirically set as 1 second \cite{Filter}.
To stabilize the PoRs at a certain distance, we set the scale factor $j$ as 2.
In fact, users do not need to know the depth of hidden objects. 
The user only needs to try to focus further, and as long as the gaze depth exceeds $w+\delta$, the hidden object is shown. Then the hidden object will guide the users' gaze to be stabilized at its depth.

\textbf{Self-Control (SC) see-through mode}.
The SC see-through mode enables the user to freely control vergence eye movement without the need for a stimulus, as shown in Fig. \ref{figure:control_mode}b, which is novel and appealing.
The user first stands $w$ meters away from the wall.
Then the user needs to perform a voluntary divergence eye movement to trigger see-through vision.
This action is completed by contracting the extraocular muscles of the eyes \cite{demer2018functional}.
The range of $w$, the distance threshold $\delta$, the time window of $\Phi\left(\cdot\right)$, the scale factor $j$, and the computation of window position $P_{\textit{\text{window}}}$ are the same as in the first mode.

\begin{figure}[t]
	\centering 
	\includegraphics[width=\columnwidth]{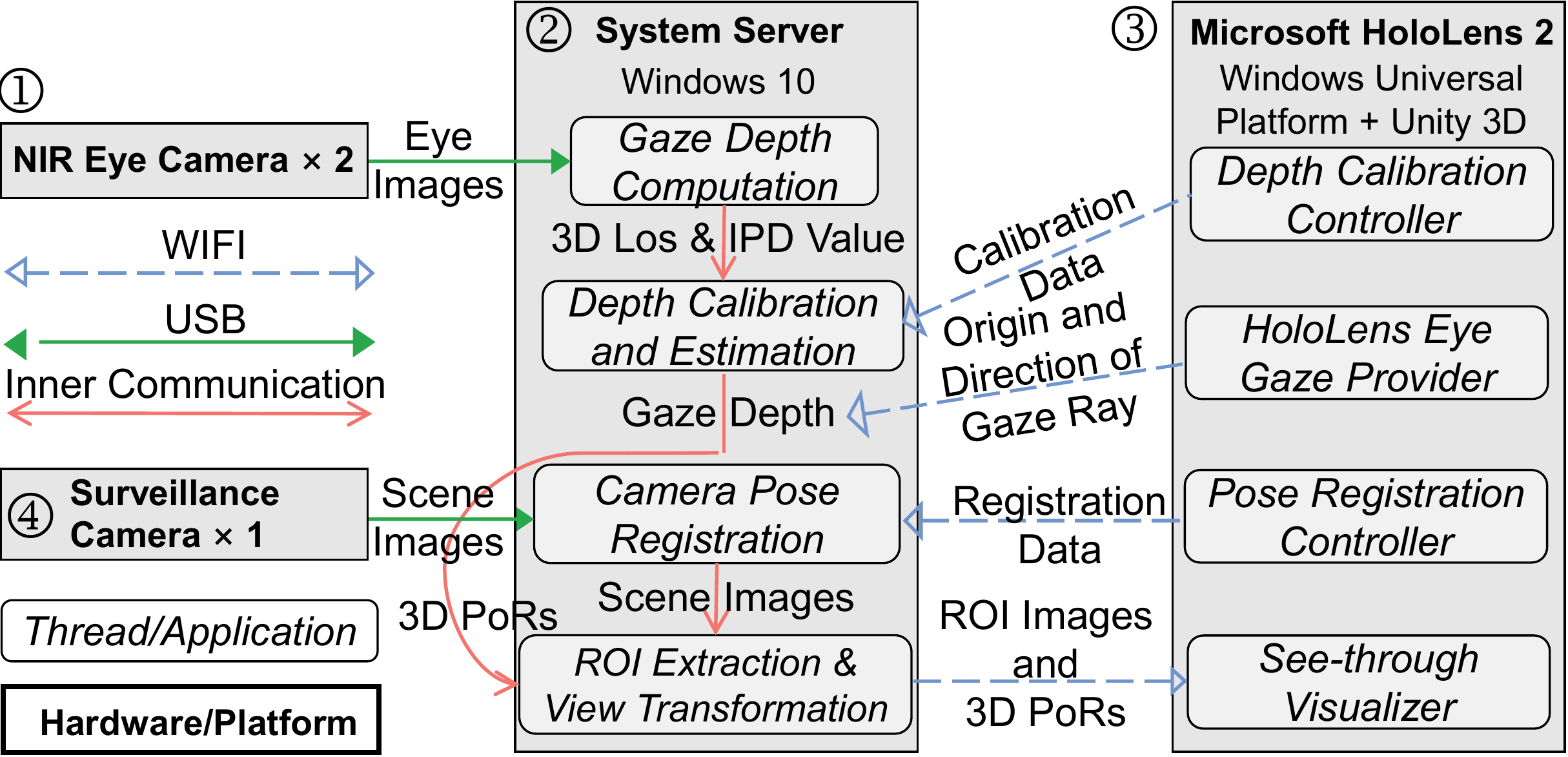}
	\caption{The system architecture of our experimental setup.
		The system includes four main hardware components:
		1) the customized NIR eye cameras,
		2) the system server,
		3) the HoloLens,
		4) the surveillance camera.
		The software components run on the system server and HoloLens.
	}
	\label{figure:System_Implementation}
\end{figure}

\subsection{Hidden Scene Capture}

The flexible control of see-through vision by gaze depth is elaborated on the last section.
We further introduce how to acquire the content of see-through vision from the hidden scene.
Here, we expect see-through vision to be natural and realistic.
For example, the user's view is consistent with physical laws, \textit{i.e.}, the presented content consistent with that the user directly
sees the scene without the occluding wall.
Besides, the window of see-through vision naturally follows the gaze direction.
To address these requirements, our solution consists of three steps:
1) To capture hidden scene, we embed a surveillance camera behind the occluding wall. 
The camera is first registered to the HoloLens coordinates.
2) We further compute the Region of Interest (ROI) of users in the HoloLens space and map the ROI into the camera space.
3) We finally perform a perspective transformation to transform the image of ROI into the user's view in HoloLens.
We illustrate these steps as follows.

\textbf{Camera Pose Registration}.
The goal of this step is to register the camera coordinates to the HoloLens coordinates.
We first manually align a virtual cuboid with a chessboard in AR and then register the camera to the HoloLens space $H$ by detecting the chessboard.
The width and length of this cuboid are equal to the size of the chessboard.
We collect 2D pixel coordinates of the chessboard in $C$ and 3D coordinates of the cuboid corners in $H$.
Finally, we use the Efficient Perspective-n-Point (EPnP) algorithm \cite{lepetit2009epnp} to compute the transformation $T$ from $H$ to $C$.

\textbf{ROI Extraction of Hidden Scene}.
We naturally control the content of our see-through vision with eye movement.
In short, we compute the ROI in the camera space $C$ and clip the image of the ROI.
Specifically, we first define the 3D PoR as the center of the user's view (ROI) in $H$, which is a rectangle plane and perpendicular to the gaze ray.
Then we compute the ROI in the camera space $C$ by using the transformation $T$.
Finally, we clip the image of ROI from the 2D image space of $C$.

\textbf{Perspective Transformation}.
The user's pose is different from the camera pose.
In practice, it causes the viewpoint difference between them. 
Therefore, to make the user's view consistent with physical laws, we map the image of ROI into the HoloLens space $H$.
We apply the perspective transformation method \cite{Perspective_Transformation} to transform them.
For the final visual effect, please refer to Section 6.

\subsection{System Implementation}

A detailed overview of our system architecture, including hardware and software implementation and data flows between them, is shown in Fig. \ref{figure:System_Implementation}.
We describe the implementation details as follows.

\begin{figure}[t]
	\centering 
	\includegraphics[width=0.8\columnwidth]{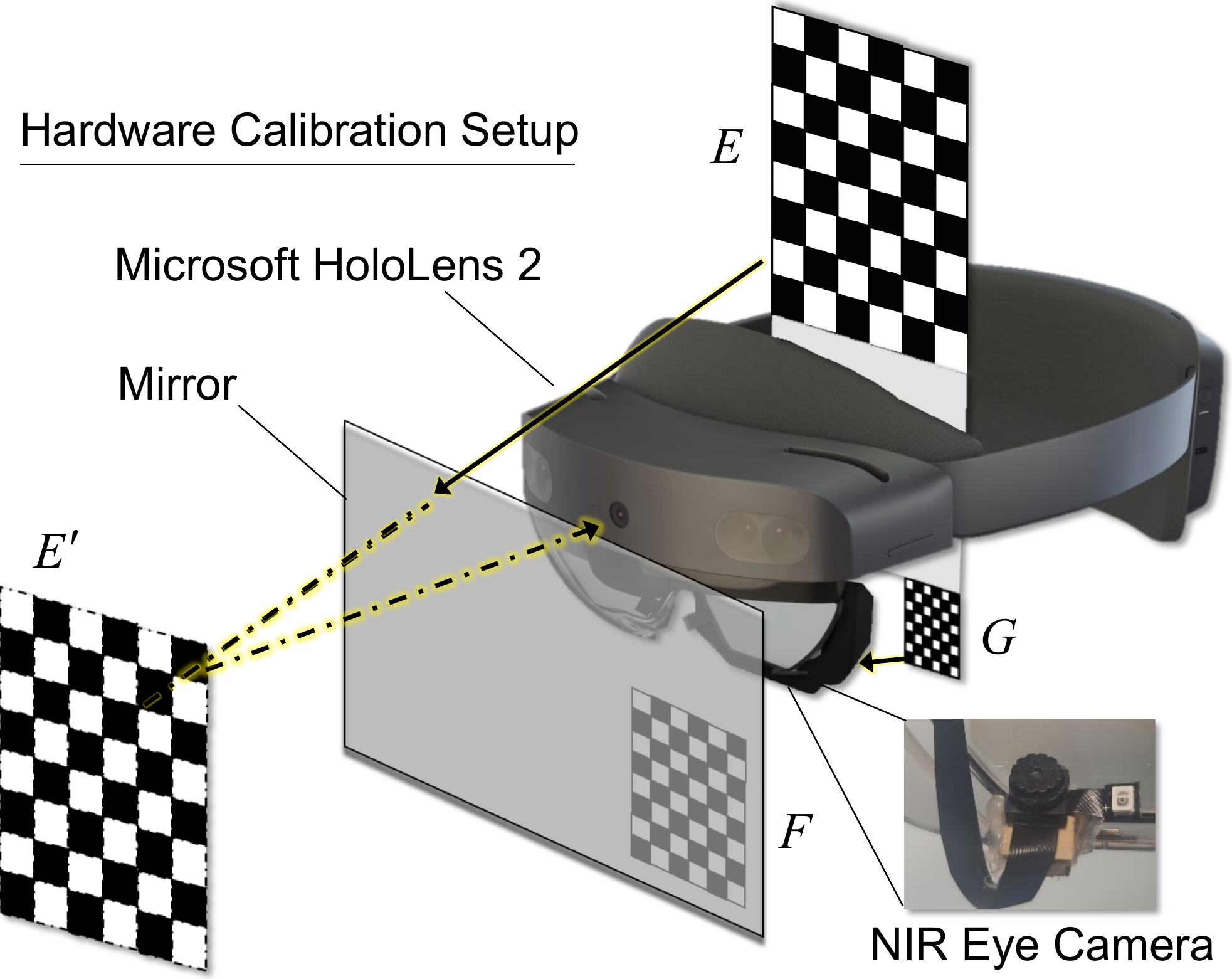}
	\caption{A hardware calibration setup is used for calibrating the transformation between the scene camera and the customized eye cameras, including the two-chessboard pattern $E$-$G$, and another chessboard $F$ attached to a mirror.}
	\label{figure:two_chessboard}
\end{figure}

\textbf{Hardware Setup}.
Our main hardware components are as follows: 
1) The customized NIR eye cameras for gaze depth estimation are shown in Fig. \ref{figure:two_chessboard}. 
The eye cameras capture near-infrared images at 30Hz with $320 \times 240$ resolution. 
2) The server uses an Intel Core i5-8500 with a 3.00Ghz CPU.
3) We use the Microsoft HoloLens 2 as the AR HMD.
4) One Logitech C9320e camera is used at 30Hz with $800 \times 600$ resolution. 

\textbf{Hardware calibration}. 
This module is used for calibrating the transformation between the scene camera and the customized NIR eye cameras.
The hardware calibration is explored by Itoh \textit{et al.} \cite{ItohK14}, who built a five-marker setup and registered these markers to a common coordinate system.
This differs from our approach in that we employ a two-chessboard pattern and another chessboard attached to a mirror, as shown in Fig. \ref{figure:two_chessboard}, which is inspired by the mirror-based extrinsic calibration \cite{6247783}.
We minimize the number of markers, which can reduce the error caused by unifying different coordinate systems.
The following describes the calibration procedure.
1) The scene camera detects the virtual image $E'$ of the chessboard $E$, and we can compute the pose of $E'$ in the scene camera coordinate system $S$.
2) The scene camera captures the chessboard $F$, and we can obtain the pose of the mirror in $S$.
3) Through the mirror symmetry, we can compute the pose of $E$ in $S$.
4) The eye camera captures the chessboard $G$ and the pose of $G$ in the eye camera coordinate system $N$ is obtained.
5) The $E$ and $G$ are coplane, and therefore they can be easily registered to a same coordinate system.
Therefore, the $S$ and $N$ can share the common coordinate system.

\textbf{Software Architecture}.
The software components and data streams between them are shown in Fig. \ref{figure:System_Implementation}.
We use the MessagePack to un/pack the data \cite{MessagePack}
and the NetMQ for network communication \cite{NetMQ}.
We use Unity 3D for visualization on the HoloLens.
The rendering rate on HoloLens is 60 fps while the frame rate on the system server is 30 fps.


\section{Quantitative Evaluation of Gaze Depth Estimation}

Gaze depth estimation is one of the most important parts of our method, and therefore we first evaluated the depth accuracy of our proposed methods, namely 3D LosI, MIPD, and PIPD, described in Section 3.2, with the Pupil Labs 3D tracker \cite{Pupil_Labs}.
We recruited 12 subjects from the campus (9 males and 3 females).
The average age of participants is 23.9 (SD = 1.55).
Three users had normal vision and nine users wore glasses.
The experiments were conducted in an AR environment.


\textbf{Design and Procedure}. 
We designed a test scenario for evaluating these methods, which is similar to the calibration scenario in Section 3.2.2.
We first introduced the experimental procedure to the participants.
Then they performed gaze depth calibration as shown in the left part of Fig. \ref{figure:depth_fitting}.
After that, participants began the test phase.
In this phase, the gaze targets will appear 18 times in a random order within the range of 0.5 to 6 m.
The size and duration of targets are the same as the calibration scenario.
We collected pupil data and gaze targets to compare these methods simultaneously.

\begin{figure}[t]
	\centering
	\includegraphics[width=0.8\columnwidth]{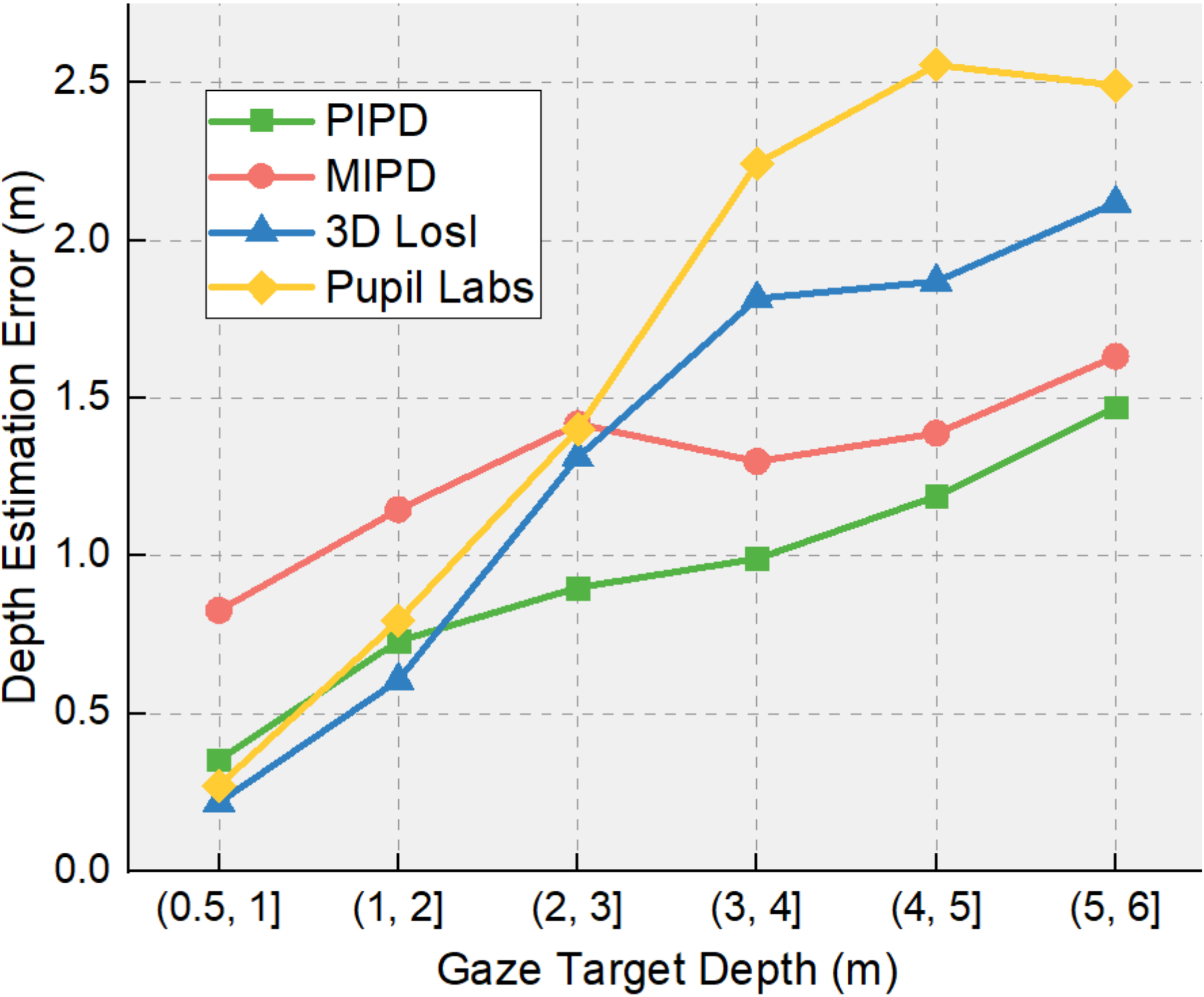}
	\caption{Mean error comparison of our gaze depth estimation methods (PIPD, MIPD and 3D LosI) with the Pupil Labs 3D tracker. 
		The standard deviation is not annotated in this figure for clear comparison. 
	}
	\label{figure:depth_error}
\end{figure}

\begin{table}[!t]
	\setlength\tabcolsep{3pt}
	\centering
	\renewcommand\arraystretch{1.3}
	\caption{The error of depth estimation (m). The first row represents the distance range. The second to fifth rows include the mean of the error and its standard deviation.}
	\begin{tabular}{c|cccccc}
		\hline
		Distance & (0.5, 1] & (1, 2] & (2, 3] & (3, 4] & (4, 5] & (5, 6] \\ \hline
		PIPD    & 0.3$\pm$0.3 & 0.7$\pm$0.5  & 0.9$\pm$0.4  & 1.0$\pm$0.5  & 1.2$\pm$0.3  & 1.5$\pm$0.5  \\ \hline
		MIPD    & 0.8$\pm$1.2 & 1.1$\pm$0.7  & 1.4$\pm$0.8  & 1.3$\pm$0.5  & 1.4$\pm$0.7  & 1.6$\pm$0.6  \\ \hline
		3D LosI    & 0.2$\pm$0.1 & 0.6$\pm$0.4  & 1.3$\pm$0.9  & 1.8$\pm$0.7  & 1.9$\pm$0.4  & 2.1$\pm$0.4  \\ \hline
		Pupil Labs    & 0.3$\pm$0.2 & 0.8$\pm$0.5  & 1.4$\pm$0.7  & 2.2$\pm$0.6  & 2.6$\pm$0.5  & 2.5$\pm$0.4  \\ \hline
	\end{tabular}
	
	\label{tab:Error}%
\end{table}%

	

	

\textbf{Results}.
We used the absolute difference between the estimated depth and the truth as an error evaluation metric.
Quantitative comparison results are shown in Fig. \ref{figure:depth_error} and Table \ref{tab:Error}, from which we make the following observations:
1) Overall, the 3D LosI achieves the best performance in the range of (0.5, 2] m (error = $0.41 \pm 0.34$ m), while the PIPD outperforms the other methods at the (2, 6] m ($1.14 \pm 0.49$ m).
2) Our 3D LosI ($1.32 \pm 0.88$ m) surpasses the Pupil Labs 3D Tracker ($1.63 \pm 1.01$ m) in all range of (0.5, 6] m.
3) We found the MIPD method has the highest error in the range of (0.5, 2] m ($0.94 \pm 1.03$ m) due to two outliers.






\textbf{Discussion}.
We discuss our results in three aspects.
1) The 3D LosI method slightly outperforms the PIPD at (0.5, 2] m.
However, the error tends to increase abruptly with a slope of 0.6 at (2, 4] m.
We argue that this is because after gaze depth exceeds 2 m, the accuracy of 3D LosI cannot meet the requirement that this method needs to correctly discriminate $1\degree$ vergence difference between 2 m and 4 m distance.
2) To overcome the above limitation, we build an optimal piecewise function for the gaze vergence control.
Specifically, if the result of PIPD is in the range of (0.5, 2] m, we use the output of 3D LosI; otherwise, we still utilize the PIPD to estimate depth.
We demonstrate that this piecewise function works efficiently for GVC techniques in the following section.
3) Our primary goal is to use the gaze vergence to perform daily indoor interaction within the middle distance, \textit{i.e.}, (0.5, 3] m, as illustrated in Section 3.3.
The gaze depth estimation error is $0.57 \pm 0.44$ m as predicted by the piecewise function.
To increase the error-tolerant rate of the GVC techniques, we use the distance threshold $\delta$ as described in Section 3.3, which is set as the sum of mean error and standard deviation at each distance.

\section{Comparisons of Interaction Modalities for See-through Vision Control}



The primary goal of this section is to evaluate and compare the Gaze-Vergence-Controlled (GVC) techniques with two common modalities in AR see-through vision.
There are few works in the literature that discuss how to control see-through vision by gaze depth in a flexible manner.
We also want to know whether the GVC techniques have advantages over other modalities.
To this end, we implement Stimulus-Guided Gaze (\textit{SGGaze}) and Self-Control Gaze (\textit{SCGaze}) and two conventional interactions, \textit{i.e.}, midair click technique (\textit{Click}) and speech-based technique (\textit{Speech}).
We propose two hypotheses:

\noindent${\text{H}_{1}}$: \textit{Controlling see-through vision has higher efficiency and usability with the GVC techniques than using Click and Speech within the middle distance.}

\noindent${\text{H}_{2}}$: \textit{Controlling see-through vision is more intuitive and attractive with the GVC techniques than using Click and Speech within the middle distance.}

\begin{figure}[t]
	\centering
	\includegraphics[width=0.9\columnwidth]{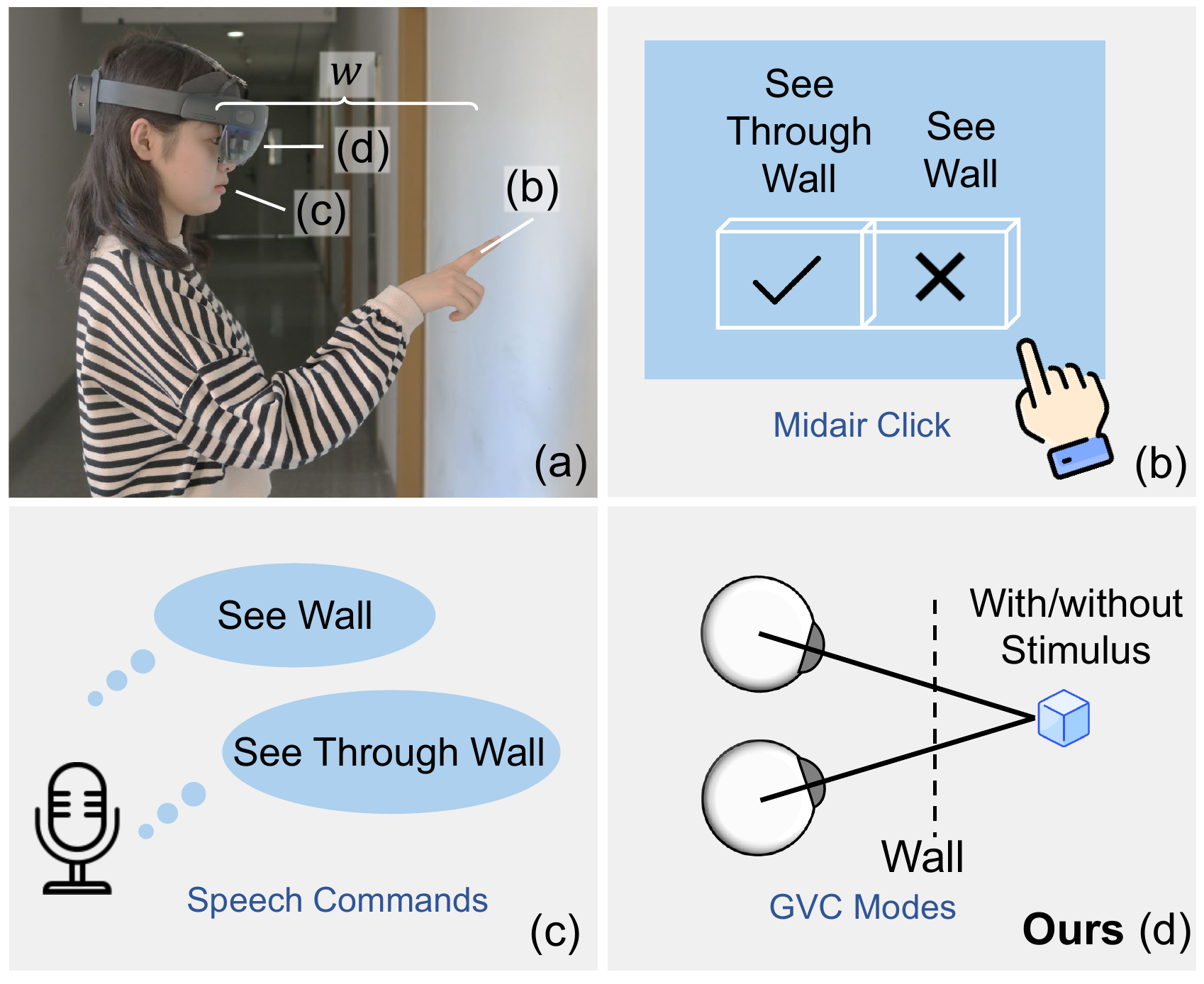}
	\caption{
		The setting of see-through vision and the illustrations of four techniques.
		(a) The user employs four interaction modalities to control the see-through vision. $w$ represents the distance between the user and the wall.
		(b) Midair click technique (Click). 
		(c) Speech-based technique (Speech). 
		(d) Stimulus-guided Gaze (SGGaze) and Self-control Gaze (SCGaze).
	}
	\label{figure:Four_Modal}
\end{figure}

\begin{figure}[t]
	\centering
	\includegraphics[width=0.95\columnwidth]{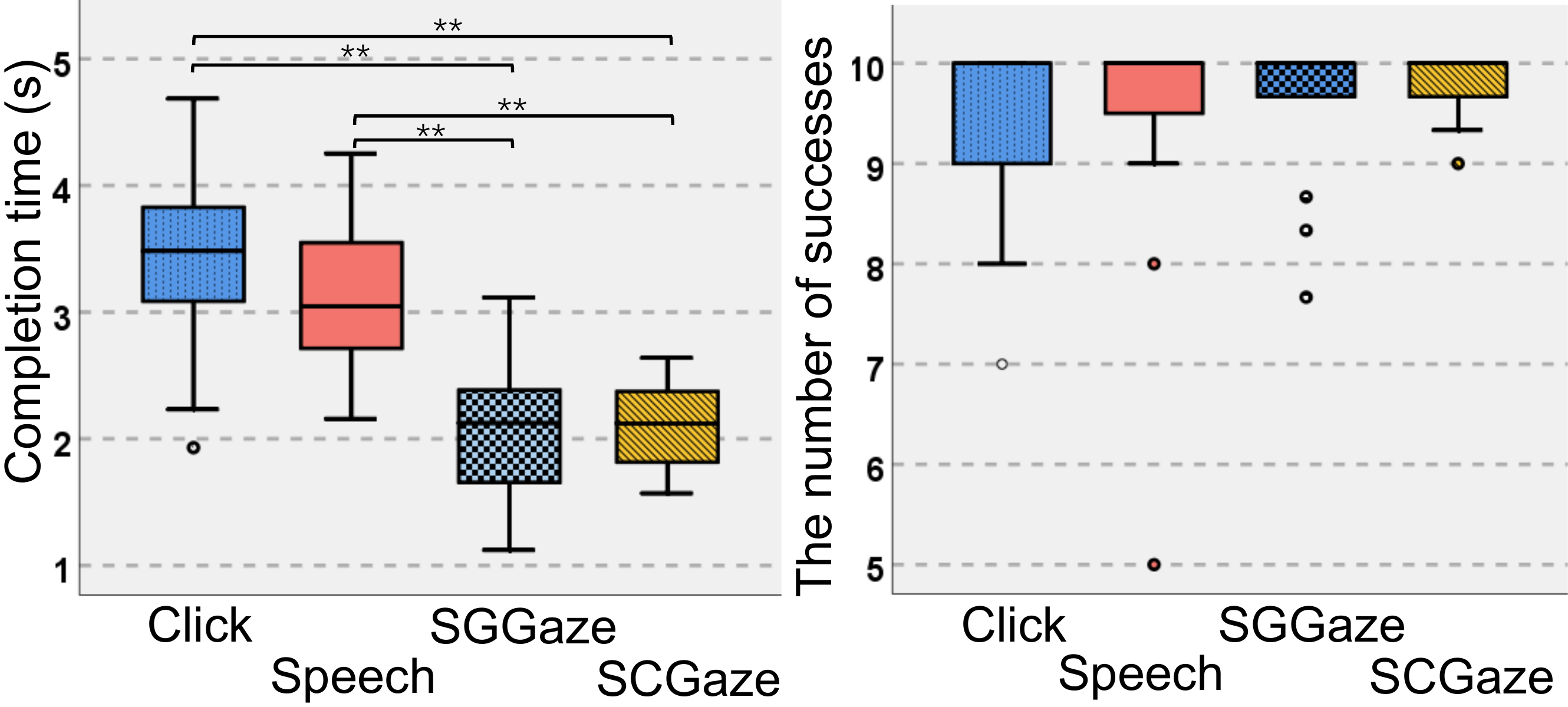}
	\caption{
		Left: Boxplots of completion time of four modalities. Right: Boxplots of the number of successes of four modalities. The
		statistical significance is labeled with ** (p \textless 0.05). Error bars mean
		standard deviations. The little colored circles indicate the outliers. There is
		no statistically significant difference in the number of successes.}
	\label{figure:objective_results}
\end{figure}

\begin{figure}[t]
	\centering
	\includegraphics[width=0.75\columnwidth]{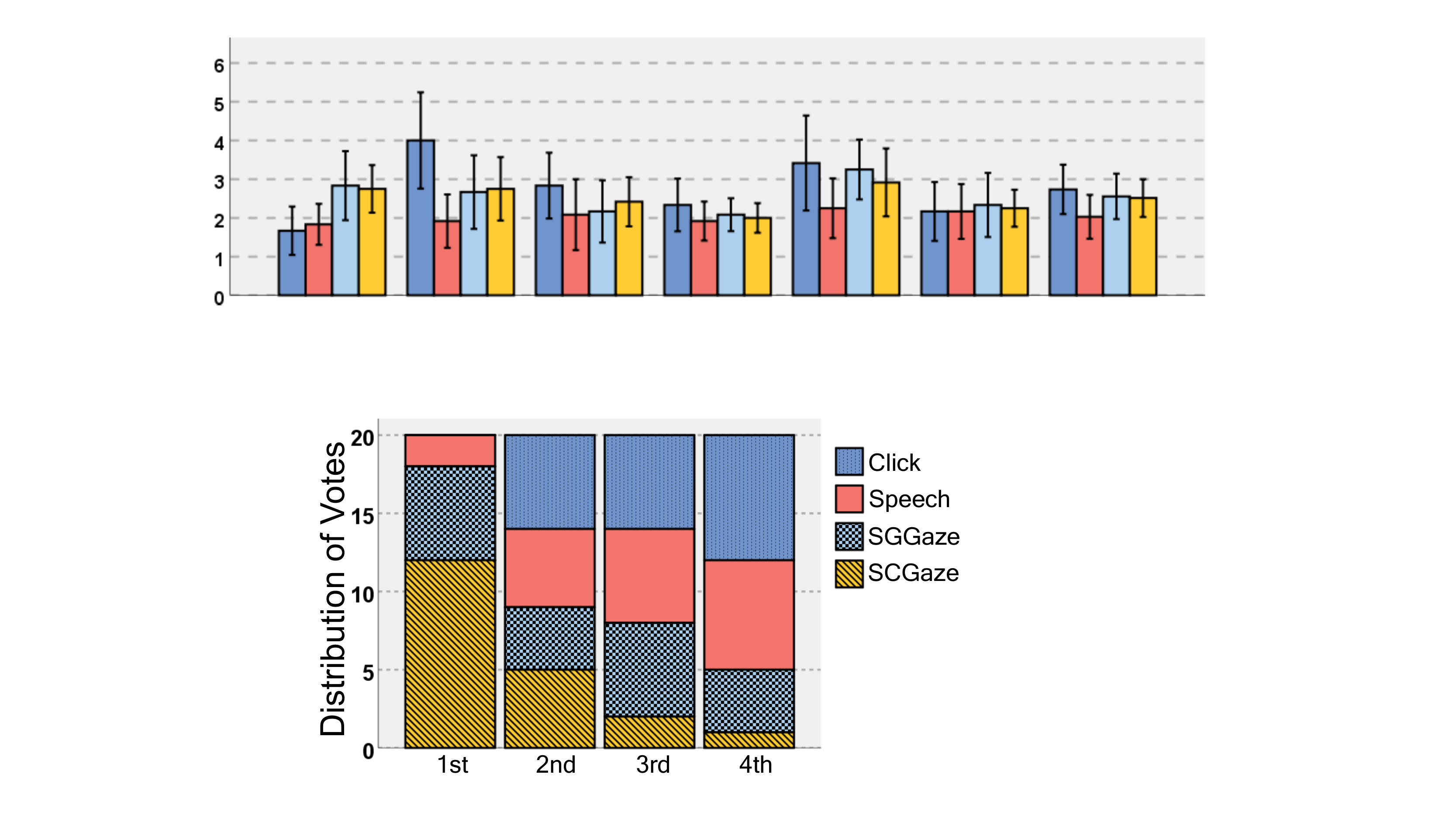}
	\caption{
		The user preference ranking of four interaction modalities.
		The \textit{SCGaze} is the most preferred bu the users.}
	\label{figure:user_preference}
\end{figure}

\subsection{Participants and Task} 

We recruited 20 subjects from the campus (12 males and 8 females).
The average age of participants is 24 (SD = 1.6).
The pre-study questionnaire with 5-point Likert scales shows the participants have low prior familiarity with AR (Mean = 2.9), the eye tracker (Mean = 2.9), medium familiarity with speech-based inputs (Mean = 3.4), and high familiarity with button-based interaction (Mean = 4.2).
All users can read and speak English fluently. 

The task requires participants to control the visualization of occluded areas according to operating commands.
In each trial, the distances between users and the wall are randomly chosen as 1, 2, or 3 m.
Specifically, the user first views the objects attached to a wall.
Then the user should do the following steps:
1) When the AR HMD displays the ``See Through Wall", the user tries to see through the wall using different modalities: clicking the ``Check" button, saying ``See Through Wall" or increasing the gaze depth. 
2) Once the system shows the ``See Wall", the participant does the opposite operations to close the see-through vision using four techniques.
After each command is successfully performed, the user needs to keep the same state for 5 seconds and wait for the next command (called the waiting state later in Section 5.5.1).
The user is required to repeat the above two steps five times.


\subsection{Interaction Modalities}


\textbf{Midair Click Technique}.
The user employs index finger to touch the ``Check" or ``Uncheck" button for controlling see-through vision, see Fig. \ref{figure:Four_Modal}b.
This technique is independent of the distance between users and the wall.
Therefore, we evaluate it at a distance of 1 m.

\textbf{Speech-based Technique}.
The participant uses the verbal command ``See Through Wall" to see the occluded regions
and says ``See Wall" to turn off the see-through vision, see Fig. \ref{figure:Four_Modal}c.
This modality is also unrelated to the distance and thus tested at a distance of 1 m.

The $SGGaze$ and $SCGaze$ techniques are implemented as described in Section 3.3, which is shown in Fig. \ref{figure:Four_Modal}d.
Our goal is to use the GVC techniques in daily indoor interaction within the middle distance, \textit{i.e.}, (0.5, 3] m.
Therefore, we set the distance as 1, 2, and 3 m.
We aim to explore whether different distances have an influence on performance.

\begin{figure*}[!t]
	\centering 
	\includegraphics[width=0.9\linewidth]{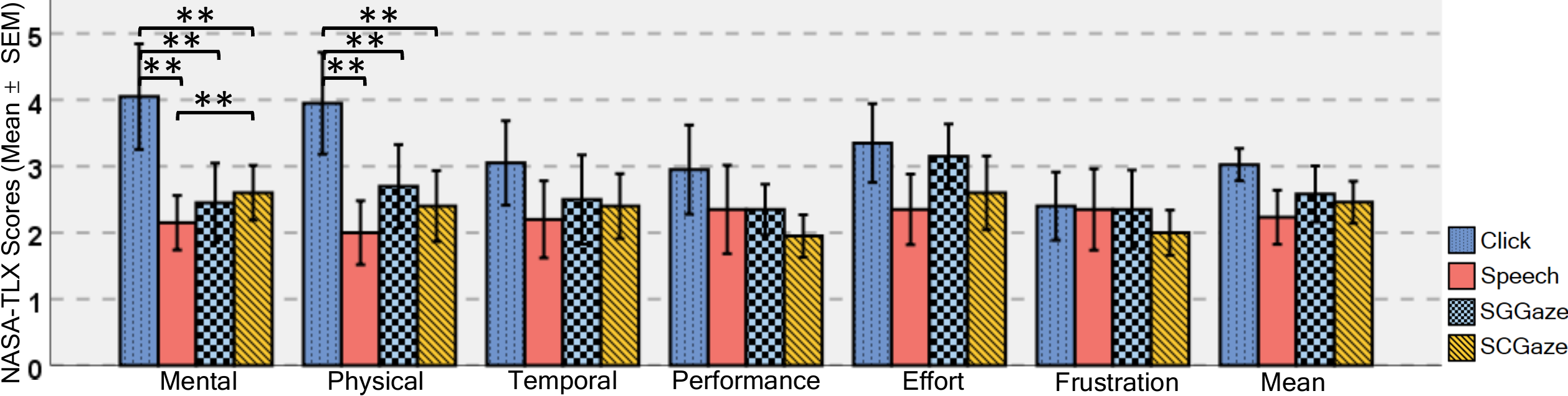}
	\caption{Bar charts of scores on the NASA-TLX questionnaire for comparing four modalities. The statistical significances are labeled with ** (p $\textless$ 0.05). Error bars mean standard deviations.}
	\label{figure:TLX}
\end{figure*}

\begin{figure*}[!t]
	\centering 
	\includegraphics[width=0.9\linewidth]{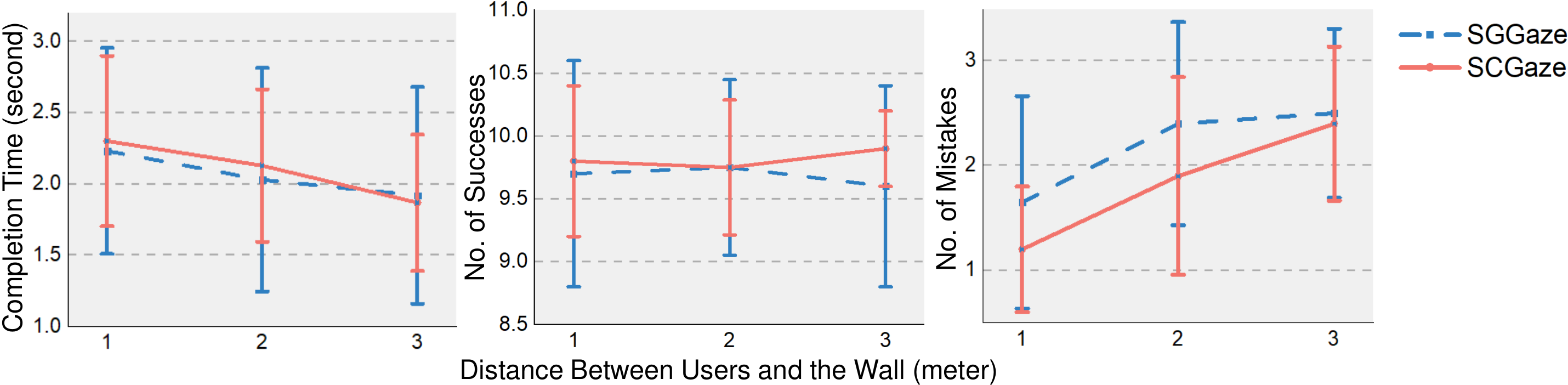}
	\caption{Line charts of GVC techniques' performance under different measurements.
		The comparison of completion time (left), the number of successes (center), and the number of mistakes (right).
		Error bars mean standard deviations.}
	\label{figure:Depth}
\end{figure*}

\subsection{Evaluation Metrics}

\textbf{Performance Measures}.
We employ three objective metrics to capture user performance: completion time, the number of successes, and the number of mistakes.
Completion time is the time elapsed between when the operating command is displayed and when see-through vision is triggered or closed correctly. 
The number of successes is the number of times that the user triggers the corresponding operations successfully in 10 seconds.
We count the number of mistakes as the number of times that the participant unintentionally triggers false commands in the waiting state.

\textbf{Subjective Measures}. Our subjective metrics describe the usability of four modalities.
After finishing the task with one technique, users fill in the NASA's Task Load Index \cite{doi:10.1177/154193120605000909} with 7-point Likert scales.
Then they answer six free-response questions to report the naturalness and frustration of each modality.
Upon the completion of all trials, they fill out a preference ranking questionnaire to rank all the modalities according to overall preference.

\subsection{Experimental Procedure}

The participants first filled in a pre-study questionnaire.
Then they began a training phase where they were given visual and auditory instructions and practiced using different modalities.
After training, they performed the experiments, including one task using four techniques and four questionnaires.
The four interaction modalities were presented in random order.
Each common modality was tested at a distance of 1 m.
Each GVC technique ran a complete process for three distances.
Users were required to rest for 30 seconds after each process to counteract the effects of fatigue.
Finally, the participants filled out the preference ranking questionnaire.
Prior to each section associated with gaze vergence, the users conducted gaze depth calibration as described in Section 3.2.1.
For a fair comparison, the ``See-through Wall" command was shown at the location of the wall, while the ``See Wall" was displayed 0.5 m in front of the see-through vision window.
Such a setting ensures that the change of gaze vergence will not happen ahead of time.
Overall, per subject performed 80 ($=(2$ techniques $\times$ 1 distance $+$ 2 techniques $\times$ 3 distances)  $\times$ 2 steps $\times$ 5 repetitions) trials.
Each experiment took around 70 min.


\begin{figure*}[t]
	\centering 
	\includegraphics[width=0.95\linewidth]{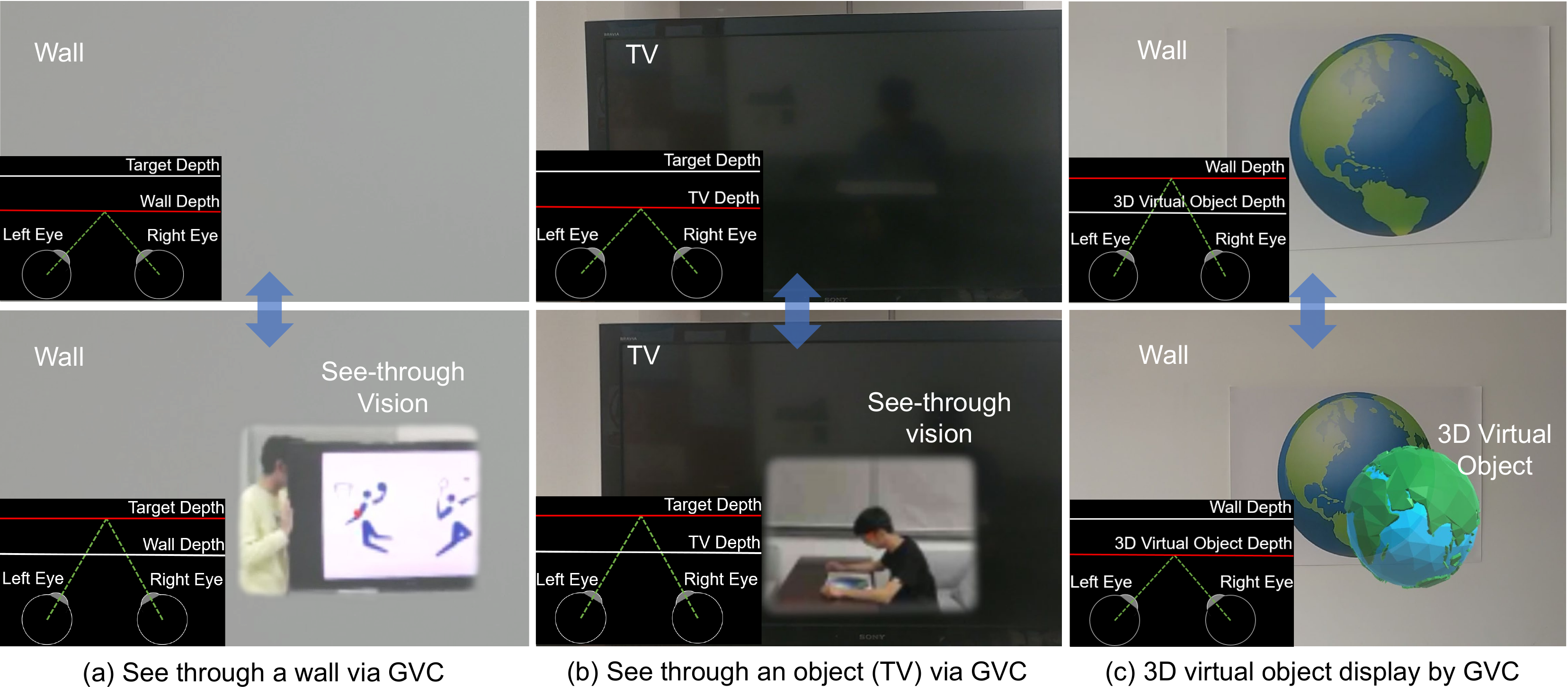}
	\caption{We present three examples to provide the implications for researchers. 
		(a) See through a wall via gaze vergence control.
		(b) See through an object via gaze vergence control.
		(c) 3D virtual object  display triggered by gaze vergence.
	}
	\label{figure:application}
\end{figure*}

\subsection{Results}

\subsubsection{Objective Evaluation Results}
\textbf{Completion Time}.
We conducted repeated-measures ANOVAs ($\alpha= 0.05$) and post hoc pairwise t-tests to judge whether the average completion time is significantly different across modalities.
For each GVC technique, we computed the average completion time of three distances.
The results are shown in the left part of Fig. \ref{figure:objective_results}. 
The statistical analysis indicated that the effect of four modalities on completion time was statistically significant ($F$(3, 57) = 37.662, $p$  $\textless$ 0.001, $\eta^{2}=0.665$). 
We found that \textit{SGGaze} was significantly faster than \textit{Click} and \textit{Speech} (p $\textless$ 0.001, 0.001). 
Besides, \textit{SCGaze} was also significantly faster than the two common modalities (p $\textless$ 0.001, 0.001). 
In order to get convincing results, we also compared the GVC techniques at distances where users spent the longest completion time with \textit{Click} and \textit{Speech}.
We found that the aforementioned significant difference still existed.

The left part of Fig. \ref{figure:Depth} shows the completion time of each GVC modality at 1, 2, and 3 m distances.
There is no significant difference between \textit{SGGaze} and \textit{SCGaze}.
We saw that the completion time of both GVC techniques gradually decreased as the distances increased.
This was expected because the change of gaze depth at the near range requires a larger rotation amplitude of the eyeballs, which results in taking longer time, while the far range did the opposite.


\textbf{The number of successes}. We performed a repeated-measures ANOVA ($\alpha= 0.05$) to identify whether the number of successes is significantly different across modalities. 
The result is shown in the right part of Fig. \ref{figure:objective_results}. 
We found it failed to reject the equality of the levels of modalities on the number of successes ($F$(3, 57) = 1.243, $p$ = 0.301, $\eta^{2}=0.061$).
The middle part of Fig. \ref{figure:Depth} plotted the average number of successes of two GVC modalities averaged across subjects.
We found that this metric was invariant to the distance.
Overall, these results indicated that users can almost finish the correct operations at the assigned time.

\textbf{The number of mistakes}.
We define the false triggering in the waiting state as the mistake mentioned before.
It did not occur to the \textit{Click} and \textit{Speech} in the waiting state.
We counted the average number of mistakes made by GVC techniques across users at three distances, as shown in the right part of Fig. \ref{figure:Depth}.
As expected, the number of mistakes increased with increasing distances.
This is because when gaze depth increases exponentially, the accuracy decreases accordingly.
We also found no significant effect of the two GVC techniques on the number of mistakes ($F$(2, 76) = 0.710, $p$ = 0.495, $\eta^{2}=0.018$) 
but the average values of $SGGaze$ are higher than those of $SCGaze$ at all three distances.
The reason for this difference
could be that the stimulus of $SGGaze$ distracted the users' attention, which was reported by some users.
We plan to minimize the transparency of the stimulus to reduce the distraction in the future.


\subsubsection{Subjective Evaluation Results}

\textbf{Task Load}.
Repeated-measures ANOVA on the NASA TLX questionnaire demonstrated that four modalities had a significant difference in \textit{mental/physical demand}.
The post hoc pairwise t-tests between the modalities were shown in Fig. \ref{figure:TLX}.
In general, the \textit{Click} achieved the highest \textit{mental/physical demand} than all the other techniques.
Users generally placed the buttons next to their hands.
According to the free response, some participants found that clicking buttons required them to look down frequently, which distracted them and degraded the user experience.
We also observed that the \textit{Speech} had lower \textit{mental demand} than the \textit{SCGaze}.
A few participants reported that they are more familiar with speech-based input than with gaze vergence control.
They felt a little nervous when using GVC techniques at the beginning.
There is no significant difference in terms of other task loads.

\textbf{User preference}. 
According to the results of the preference ranking questionnaire, the \textit{SCGaze} is the most preferred by the users, as shown in Fig. \ref{figure:user_preference}. 60$\%$ of participants believed \textit{SCGaze} ranked first in terms of preference, 30$\%$ of users preferred \textit{SGGaze} the most, and two participants liked the speech-based technique the most.

\subsection{Discussion}

In this section, we discuss and summarize the results for validating the hypotheses.
%


\noindent${\text{H}_{1}}$: \textit{Controlling see-through vision has higher efficiency and usability with the GVC techniques than using Click and Speech within the middle distance.}

%

Our results supported this hypothesis.
In terms of speed, \textit{SGGaze} outperformed \textit{Click} and \textit{Speech}; \textit{SCGaze} was also superior to both common techniques. 
Besides, user's feedback also supposed that ``Controlling see-through vision by looking closer or far away rarely requires response time" (P11).
For accuracy, although the GVC techniques occasionally occurred with false triggering caused by the accuracy of depth estimation, the number of successes was still not affected.

In terms of usability, we thought that using gaze vergence to control see-through vision was convenient and easy to use.
``After simple training, it is relatively simple and has no mental fatigue." (P2)
``I feel relaxed using it." (P7)
Users reported that ``I feel arm fatigue after \textit{Click}" (P3).
Some participants claimed that `` the speech command needs to speak aloud to trigger the switch, which is not convenient in a quiet space" (P4). 
We believed that the GVC techniques tackled the limitations of \textit{Click} and \textit{Speech} and improved the user experience. 
It freed both hands and users did not need to look away. 
It can also be done without making sounds. 
P6 and P10 had similar feelings. The above analysis accounts for the superior performance of GVC techniques.

\noindent${\text{H}_{2}}$: \textit{Controlling see-through vision is more intuitive and attractive with the GVC techniques than using Click and Speech within the middle distance.}

Our results supported this hypothesis. 
We validated it in two aspects.
1) With regard to user preference, 60$\%$ of users preferred \textit{SCGaze} and 30$\%$ of participants ranked \textit{SGGaze} first. 
Most of the participants found \textit{SGGaze} and \textit{SCGaze} to be enjoyable, 
\textit{e.g.}, ``It is amazing. I have been looking in the same direction, but the change of vergence can convey a signal of seeing through the wall, which is a novel experience for me" (P16). 
2) In terms of naturalness, the GVC techniques take advantage of our viewing habit, as when we want to see through something, it is physically associated with our gaze depth/vergence, and therefore should be naturally controlled by the eyes.
In contrast, \textit{Click} needs to interrupt the user experience and ask them to look down to press a button.
\textit{Speech} requires the participants to repeat boring commands.
P4 and P7 also expressed similar opinions.
To summarize, we believed that the gaze-vergence-controlled see-through vision is more appealing and intuitive than the common interaction modalities.

\section{Examples For Gaze Vergence Control}

In this section, we demonstrate the GVC technique with four example applications, which can give insights and implications for designers. 
For more details, please refer to the supplemental video.

%
\textbf{See Through a Wall via Gaze Vergence Control}.
We show how to see through an office wall using the proposed \textit{SCGaze} technique, as shown in Fig. \ref{figure:application}a.
The user is immersed in the occluded environment with a first-person view and naturally controls the see-through vision with his eyes.
In this example, one surveillance camera is attached to the inner wall of an office.
The user stands 1 meter away from the wall.
When the user fixates on the wall, no see-through vision is activated.    
When gaze vergence reaches the target depth, see-through vision is triggered. 
The user sees the television and the moving scene through see-through vision.
The user's view is consistent with physical laws.

\textbf{See Through an Object via Gaze Vergence Control}.
The second example enables the user to see through daily indoor objects, \textit{e.g.}, a television, with the \textit{SCGaze} modality, as shown in Fig. \ref{figure:application}b.
This indicates that we can naturally make daily life objects invisible and thus expand our vision.
In this example, the participant can see the person through the television.
The user is 2 m away from the television and 4.5 m away from the person.
When the gaze vergence reaches the depth of the person, the see-through vision is activated.
The vision window also helps to keep the eyes focused at a certain distance.
Then the user's gaze is put on the television and there is no see-through effect.

\textbf{3D Virtual Object Display Triggered by Gaze Vergence}.
Our third example shows that the GVC technique can control the display of additional information about a real object of interest.
This is potentially an effective channel in maintaining or learning scenarios, \textit{e.g.}, opening the menu when both hands are occupied \cite{DBLP:journals/tvcg/KimBBDW18, PiumsomboonDELL17}.
We implement a 3D virtual earth display triggered by gaze vergence, as shown in Fig. \ref{figure:application}c. 
In this example, we do not see through but rather see in front.
When the gaze target is on the wall, the user sees a real poster, which is located 2 m away from him.
Then the user controls the gaze vergence to fixate 0.5 m in front of the wall.
As a result, a 3D virtual earth is shown at the fixation position. 

\textbf{See Through Multiple Occluding Objects via Gaze Vergence Control}. 
Our final example shows that the user can see through multiple occluding objects with the vergence control and switch between four layers of different depths.
We set 4 layers in this demo: (0, 0.6] m centered at 0.3 m, (0.6, 1.4] m @ 1.0 m, (1.4, 3.0] m @ 2.2 m and (3.0 m, +$\infty$) @ 4.5 m.
When the gaze vergence reaches the depth of each layer, the image of each layer is shown in turn.
When the user's gaze depth decreases,  the see-through vision switches to the front layers and eventually comes back to the first layer.
Please see our supplemental video for more details.

\section{Design Implications}





Based on the results and observations of our study,
we derive a set of guidelines and implications for the design of gaze-vergence controlled techniques in AR.
\vspace{-0.1cm}
\begin{itemize}[leftmargin= 9 pt, itemsep = -2 pt]
	
	\item 
	Our results demonstrate that the proposed 3D LosI and PIPD methods perform differently in the range of (0.5, 6] m.
	We suggest that for the gaze depth estimation in the range of (0.5, 2] m, the Gaze-Vergence-Controlled (GVC) techniques can utilize the 3D LosI method.
	For a depth beyond 2 m, the PIPD method can be used.
	
	\item Providing an error-tolerant design for the GVC see-through vision can increase its robustness.
	We suggest setting a distance threshold for the control.
	Considering the depth estimation error, we use the sum of mean error and standard deviation as the threshold at each distance.
	Besides, we recommend using a filter function for smoothing depth values.
	
	
	\item 
	The window of see-through vision should be fixed at a certain distance, which helps prevent the window drifting due to gaze error and stabilizes user's gaze depth to avoid frequent gaze adjustment. Meanwhile, the fixed window depth also avoids causing visual fatigue. 
	
	\item 
	During our experiments, we found that divergence movement can in fact be successfully performed but cannot last long. This is because the fixation points fall behind a wall in an instant, but they cannot be fixed without stimulus and thus come back to the wall plane quickly. Fortunately, the mechanism of our SCGaze can help avoid this problem. Our system can rapidly capture the instant change of vergence and activate see-through vision. The window of see-through vision can serve as a stimulus to help stabilize the user's vergence.
	
\end{itemize}

\section{System Limitations and Future Work}

The GVC techniques have higher efficiency and usability than using $Click$ and $Speech$ for controlling the see-through vision within the middle distance.
However, when the distance exceeds 3 m, it is difficult to discriminate the vergence difference as described in Section 4.
Therefore, for long-distance interaction ($\textgreater 3 $ m), we can use the modality independent of the distance, \textit{e.g.}, speech-based technique.



In the future, we will design and implement a shared see-through vision between multiple users controlled by gaze vergence.
In the current setting, we embedded a surveillance camera behind the wall to achieve see-through vision.
For different users staying in adjacent rooms, we plan to enable them to wear a HoloLens 2.
We can thus obtain images from the scene camera of each HoloLens.
Each user can use gaze depth to trigger the see-through vision.
The images are captured by different devices in adjacent rooms.

\section{Conclusion}
In this work, we proposed using the gaze-vergence-controlled see-though vision in AR.
We first built a gaze tracking module with two infrared cameras and assembled it into the Microsoft HoloLens 2.
With our gaze depth estimation algorithm, the user' gaze depth can be computed from gaze vergence and used to control the see-through vision.
We evaluated the efficiency and usability of four interaction techniques.
Experimental results demonstrated that gaze depth estimation is efficient and accurate.
It showed that the GVC techniques are superior in terms of efficiency and more preferred by users.
We also showed four example applications of GVC see-through vision.

\bibliographystyle{abbrv-doi}

\bibliography{template}
\end{document}